\documentclass[letterpaper, 10 pt,conference]{ieeeconf}
\IEEEoverridecommandlockouts
\usepackage{algorithmic}
\usepackage{array}
\usepackage{amsmath}
\newcommand{\bs}[1]{\boldsymbol{#1}}
\usepackage{float}
\usepackage{cite}
\usepackage{subcaption}
\usepackage{balance}
\usepackage{caption}
\usepackage[hidelinks]{hyperref}

\usepackage[pdftex]{graphicx}
\graphicspath{{../pdf/}{../jpeg/}}
\DeclareGraphicsExtensions{.pdf,.jpeg,.png}
\usepackage{epstopdf}

\usepackage{geometry}
\title{\LARGE \bf
Design and Control of A Tilt-Rotor Tailsitter Aircraft \\ with Pivoting VTOL Capability}

\author{Ziqing Ma$^{1}$, Ewoud J.J. Smeur$^1$ and Guido C.H.E. de Croon$^{1}$% <-this % stops a space
\thanks{$^{1}$All authors are with MAVLab, Faculty of Aerospace Engineering, Delft University of Technology, Delft, Netherlands. (emails: {\tt\small{Z.Ma@tudelft.nl; e.j.j.smeur@tudelft.nl; g.c.h.e.deCroon@tudelft.nl}}).}%
}
\begin{document}
\maketitle
\thispagestyle{empty}
\pagestyle{empty}

\renewcommand{\thefootnote}{}
\footnotetext{Video: \href{https://youtu.be/8O7NlUHBlJ4}{https://youtu.be/8O7NlUHBlJ4}}
\renewcommand{\thefootnote}{\arabic{footnote}}
%%%%%%%%%%%%%%%%%%%%%%%%%%%%%%%%%%%%%%%%%%%%%%%%%%%%%%%%%%%%%%%%%%%%%%%%%%%%%%%%
\begin{abstract}
Tailsitter aircraft attract considerable interest due to their capabilities of both agile hover and high speed forward flight. However, traditional tailsitters that use aerodynamic control surfaces face the challenge of limited control effectiveness and associated actuator saturation during vertical flight and transitions. Conversely, tailsitters relying solely on tilting rotors have the drawback of insufficient roll control authority in forward flight. This paper proposes a tilt-rotor tailsitter aircraft with both elevons and tilting rotors as a promising solution.
%The drone integrates both tilt rotors and flaps to enhance control authority across all flight regimes.
By implementing a cascaded weighted least squares (WLS) based incremental nonlinear dynamic inversion (INDI) controller, the drone successfully achieved autonomous waypoint tracking in outdoor experiments at a cruise airspeed of 16 m/s, including transitions between forward flight and hover without actuator saturation. Wind tunnel experiments confirm improved roll control compared to tilt-rotor-only configurations, while comparative outdoor flight tests highlight the vehicle's superior control over elevon-only designs during critical phases such as vertical descent and transitions. Finally, we also show that the tilt-rotors allow for an autonomous takeoff and landing with a unique pivoting capability that demonstrates stability and robustness under wind disturbances. 
\end{abstract}

%%%%%%%%%%%%%%%%%%%%%%%%%%%%%%%%%%%%%%%%%%%%%%%%%%%%%%%%%%%%%%%%%%%%%%%%%%%%%%%%
\section{INTRODUCTION}
In the field of robotics, vertical take-off and landing (VTOL) aircraft have gained increasing attention as they combine the high-speed, long-range capabilities of fixed-wing aircraft with the precise hovering and vertical take-off/landing abilities of rotorcraft. This will make them invaluable in various robotic applications such as aerial photogrammetry, environmental monitoring, and search-and-rescue operations. Compared to  other categories of VTOL aircraft like quadplanes\cite{karssies2022extended}, tilt-rotors\cite{chen2017control} (fixed-wing), tilt-wings\cite{hartmann2017unified}, tailsitters stand out for their unique transitioning between flight modes by pitching the entire body. This simplicity in mechanical design, along with their operational flexibility, positions tailsitters as a promising solution for complex missions requiring rapid transitions and sustained high-speed flight.
\begin{figure}
	\centering
 \includegraphics[width=7.2cm]{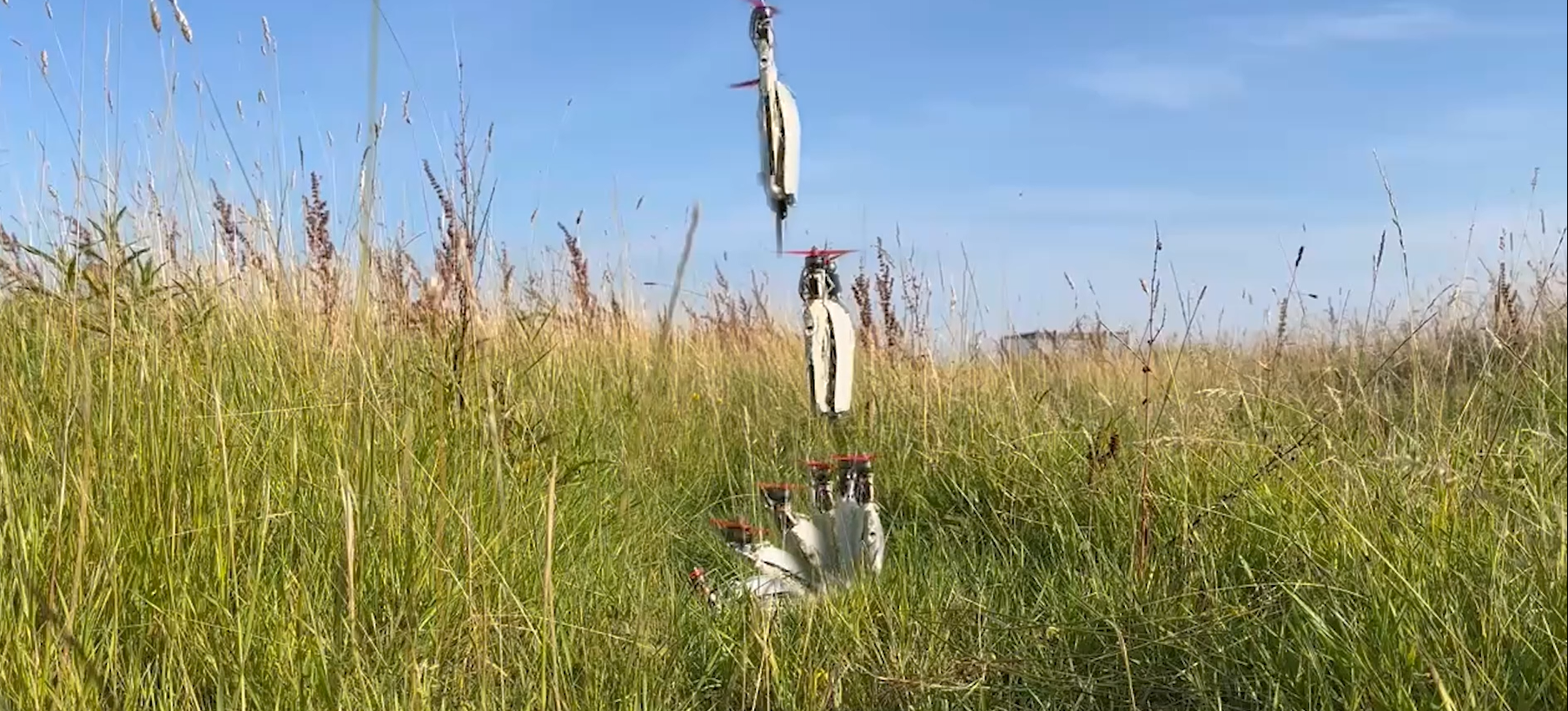}
 \includegraphics[width=7.2cm]{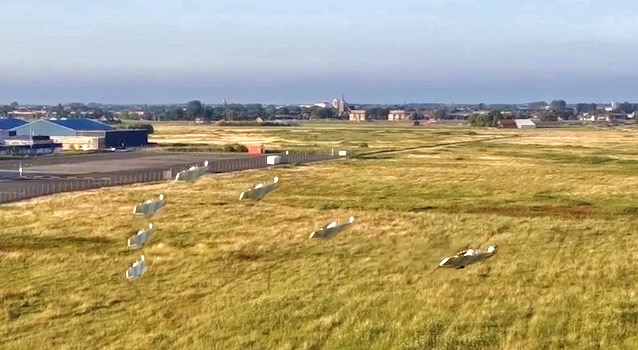}
 \caption{(Top) TRE-tailsitter executing a pivot takeoff, where it pitches up steadily off the ground by pivoting around its tail by thrust vectoring. (Bottom) Transition from hover to forward flight, captured at 0.1-second intervals.}
	\label{fig:overview}
 \vspace{-10pt}
\end{figure} 

Despite promising advancements, tailsitters face two significant challenges that hinder their broader application. The first challenge lies in robust vertical takeoff and landing. Conventional VTOL tailsitters require an upright stand on the ground before takeoff, necessitating structural elements such as additional landing gears\cite{zhong2021transition},\cite{islas2022design},\cite{anuar2024preliminary},\cite{tal2021global}. These structures, while mechanically simple, add extra weight and drag and restrict the operational flexibility of the tailsitter, especially in uneven terrain or windy conditions. For example, if a tailsitter with landing gears tips over, it will require human intervention before a new take-off is possible. This reliance on precise ground placement and human intervention reduces the autonomy and deployment flexibility of the system.

A tilt-rotor mechanism presents a promising alternative by enabling takeoff from a level resting position, as presented in Ardupilot’s Thrust
Vectored Belly Sitter project and \cite{wang2024vectored}. However, \cite{wang2024vectored} utilizes thrust vectoring solely to lift the aircraft off the ground, while exhibiting a pitch overshoot of approximately 50 degrees and significant horizontal drift.
Additionally, for the landing phase, the drone is manually caught by hand at low altitude, posing safety risks and operational constraints.
%Overall, a novel approach of autonomous vertical takeoff and landing is imperative, to enhance robustness in challenging environments, such as rough terrain or wind disturbances.

The second challenge involves controllability and actuator saturation. Tailsitters with only aerodynamic control surfaces that function like elevons (E-tailsitters) often experience limited control effectiveness at low airspeeds, leading to elevon saturation during transitions\cite{Smeur2020}. This is because the elevon effectiveness benefits from strong velocity-induced airflow at high airspeeds. The effectiveness diminishes during vertical flight or transitions, where airflow over the elevons becomes reduced or even reversed during descent, further reducing control effectiveness \cite{ma2022wind}. To address this, tailsitters using only tilt-rotors for control moment generation (TR-tailsitters) have been developed, featuring dual tilt rotors for thrust vectoring control\cite{Nogar2018}, \cite{lovell2023attitude}. Though the tilt-rotor design guarantees pitch control authority in vertical and transition phases, it suffers from reduced roll control effectiveness in forward flight due to the wing-propeller interaction. To tackle this, a tailsitter combining control surfaces with tilt rotors (TRE-tailsitter) presents a promising solution by combining the effective pitch control of tilting rotors in hover with the stable roll control provided by elevon deflections during the forward flight.

For a TRE-tailsitter, the priority is ensuring stable control throughout the flight envelope. Regarding the E-tailsitter control, substantial studies have been conducted. Compared to separate control strategies for each flight phase as presented in \cite{oosedo2017optimal}, \cite{lyu2017hierarchical}, a global control law without switching between flight modes is advantageous for rapid and smooth transitions. In \cite{ritz2017global}, a global cascaded proportional–integral–derivative (PID) controller is introduced and validated by indoor trajectory tracking tests, albeit with significant tracking errors in the pitch angle. Differential flatness has also been explored for the global control of tailsitters, as presented in\cite{tal2021global} and \cite{tal2023aerobatic}, where agile and aerobatic maneuvers are performed in indoor tests. In \cite{lu2024trajectory}, a differential flatness based model predictive controller (MPC) is also developed by establishing a high-fidelity aerodynamic model to attain aggressive flights of tail-sitters.
Nevertheless, the differential flatness property doesn't apply to the TRE-tailsitter proposed in this paper.
%which combines thrust vectoring and flap deflection for control moment generation.
In \cite{mcintosh2024aerodynamic}, a position control architecture is employed for a simplified tailsitter model in simulation, integrating a nonlinear dynamic inversion (NDI) attitude controller, which is sensitive to model inaccuracies. Alternatively, in \cite{Smeur2020}, a cascaded sensor-based incremental nonlinear dynamic inversion (INDI) controller is implemented for guidance and control of an E-tailsitter. Given the integration of tilting rotors and aerodynamic control surfaces, establishing an accurate model for the proposed TRE-tailsitter demands considerable efforts. Therefore, a control method with reduced model dependency is favored for the proposed TRE-tailsitter.

The \textbf{main contributions} of this work are as follows: 
\vspace{-2.4pt}
\begin{itemize}
\setlength{\itemsep}{0pt}
    \item Autonomous guidance and control of a TRE-tailsitter, guaranteeing control authority across the full flight envelope.
    
    \item Pivoting takeoff and landing approach as presented in Fig. \ref{fig:overview}, enabling stable takeoff and landing in windy conditions, with robustness validated through indoor and outdoor tests.
    
    \item Experimental comparison: Wind tunnel and flight tests confirm the TRE-tailsitter’s improved roll control over TR-tailsitters during the forward flight, and superior performance over E-tailsitters in descent and transitions.
\end{itemize}
The paper is structured as follows: Section II introduces the TRE-tailsitter design. Section III details the pivoting VTOL and in-flight attitude controllers, while Section IV outlines velocity control and guidance. Section V presents outdoor autonomous flight test results across the full flight envelope, and Section VI discusses pivot controller robustness and comparative fight tests. Section VII concludes the paper.

\section{AIRCRAFT DESIGN}
This section presents the design of the TRE-tailsitter aircraft, with a focus on optimizing both aerodynamic performance and control moment generation capabilities.

\subsection{Platform Configuration}
\begin{figure}
	\centering
	\includegraphics[width=0.6\linewidth]{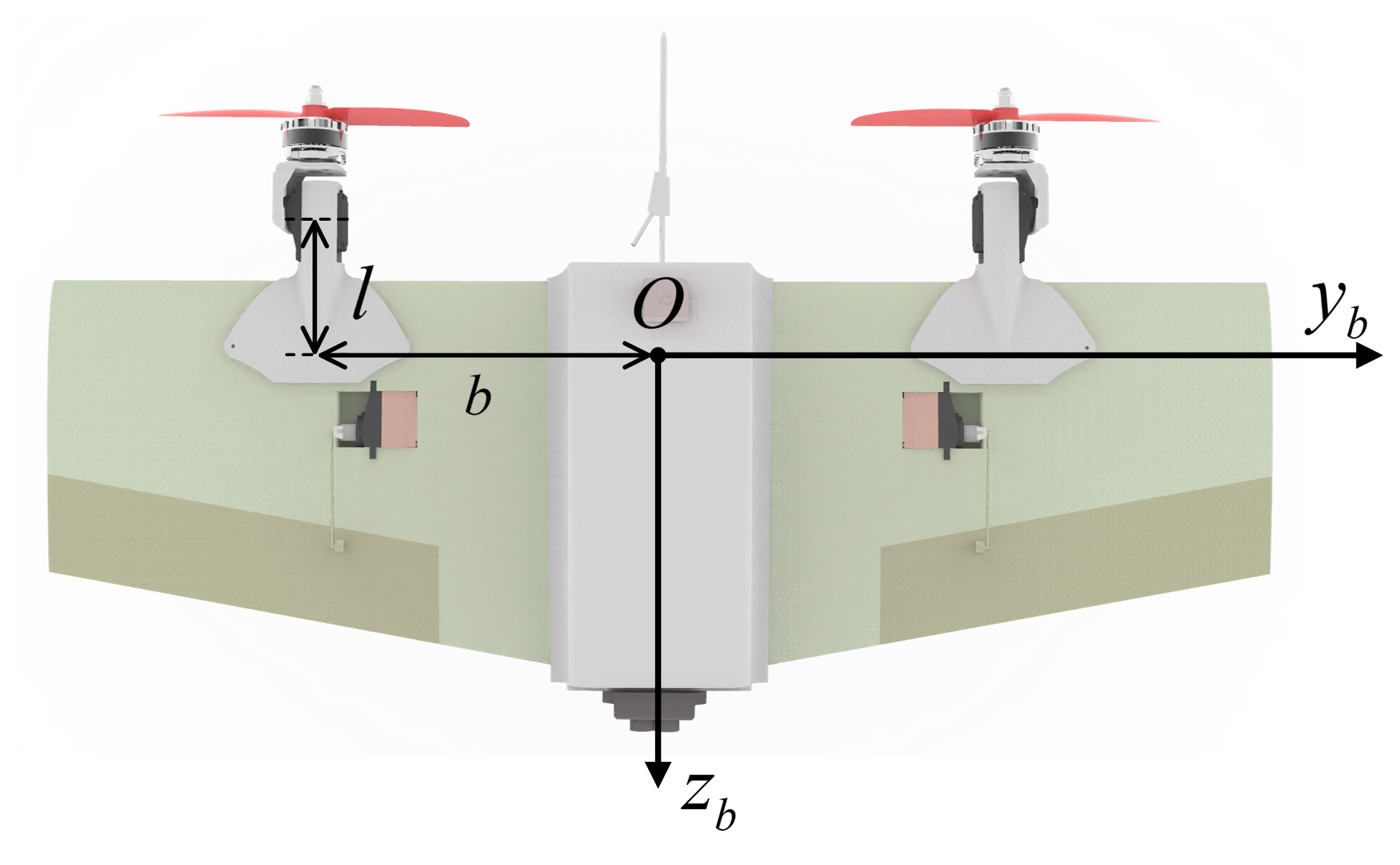}
	\caption{The top view of the TRE-tailsitter.}
	\label{fig:drone}
\end{figure}

Fig. \ref{fig:drone} shows a top view of the designed and built TRE-tailsitter, which includes a pair of elevons and dual independent tilt rotors mounted at the leading edge of the vehicle. Specifically, the tilt angles of the left and right rotors are denoted by $\delta_{TL}$ and $\delta_{TR}$, left and right elevon deflections by $\delta_{EL}$ and $\delta_{ER}$, with upward rotor tilt/elevon deflection considered positive, and a maximum tilt/deflection angle $\delta_{\max}$ of $63^\circ$. The thrusts produced by the left and right motors, $T_L$ and $T_R$. The body axis system
%for the TRE-tailsitter
is depicted in Fig. \ref{fig:drone}, following the right-hand rule. To avoid singularities at $\pm90^{\circ}$ pitch angle, the ZXY Euler rotation sequence is adopted throughout this paper.

\vspace{-10pt}
\begin{figure}
	\centering
	\includegraphics[width=0.9\linewidth]{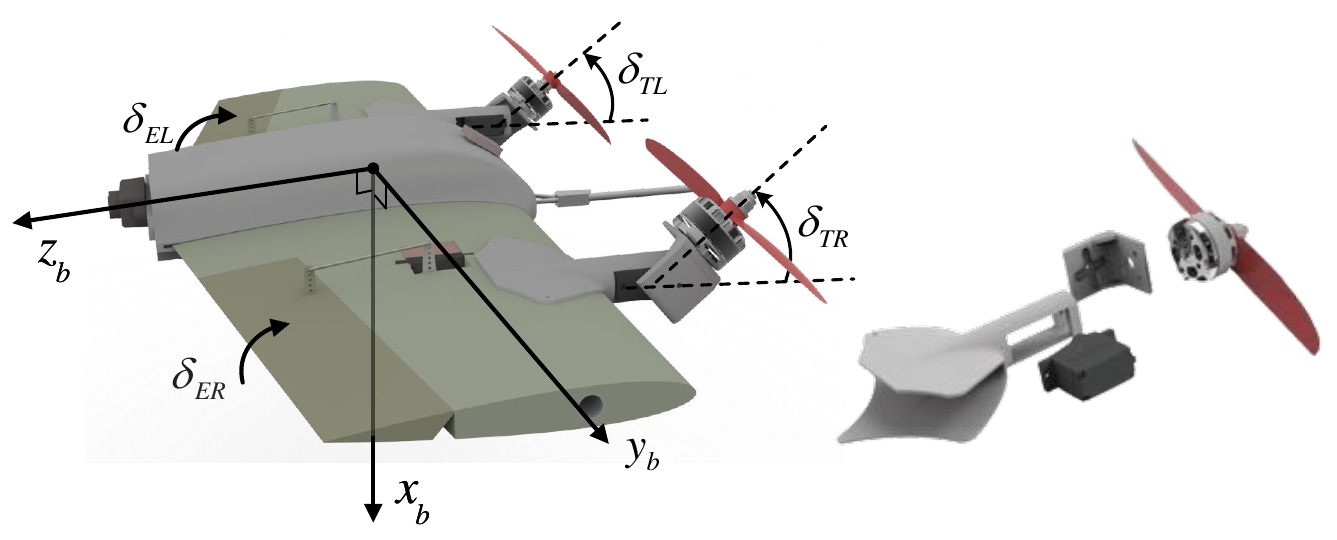}
	\caption{Tilt rotor configuration with positive rotor tilt and elevon deflection defined.}
	\label{fig:tilt}
\end{figure}
\vspace{14pt}
\begin{figure*}[h]
	\begin{minipage}[t]{0.33\linewidth}
		\begin{center}
			\includegraphics[width=\linewidth]{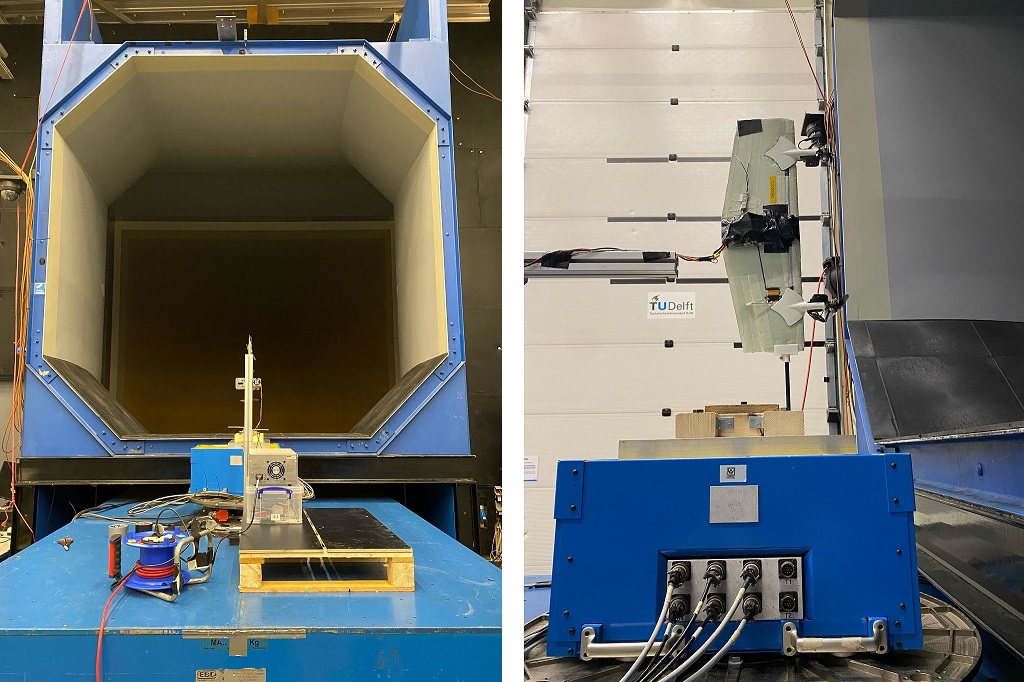}
			\caption{The front and side views for the wind tunnel test setup, where the aircraft faces against a wind airflow of 15 m/s, representing the forward flight phase.}
			\label{fig:wind}
		\end{center}
	\end{minipage}
    \hspace{0.2cm}
	\begin{minipage}[t]{0.64\linewidth}
		\begin{center}
			\includegraphics[width=0.49\linewidth]{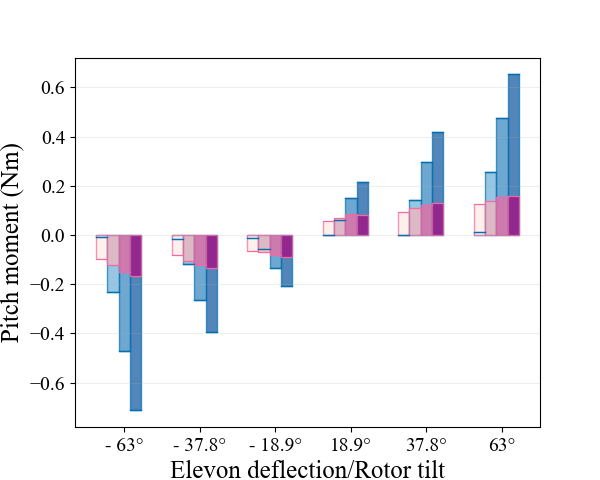}
			\hspace{-0.5cm}
			\includegraphics[width=0.49\linewidth]{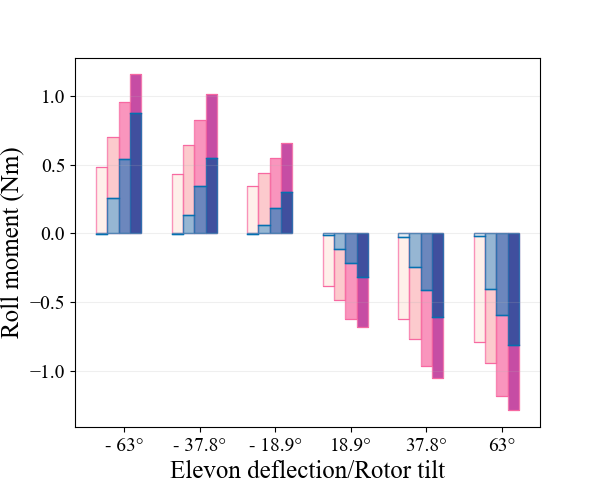}
                \includegraphics[width=0.9\linewidth]{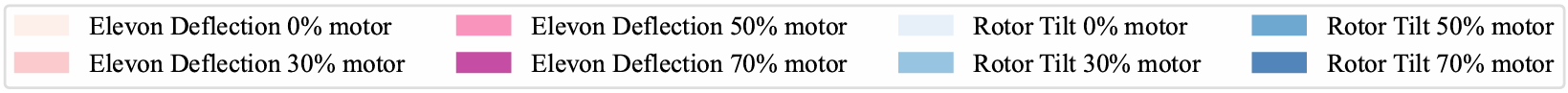}
			\caption{Pitch and roll moments generated by either elevon deflection or rotor tilt at $15$ m/s airspeed, with only one side rotor tilted or elevon deflected, under the assumption that the left and right actuators have symmetrical effects.}
			\label{fig:moment}
		\end{center}
	\end{minipage}
    \vspace{-15pt}
\end{figure*}

\vspace{-0.05cm}
The TRE-tailsitter features a wing with a NACA0012 airfoil, without a fuselage, consisting of three sections with a total wingspan of $0.5m$ and a wing area of $0.071m^2$. The center section is 0.06 m wide with a constant chord of $0.16m$, while the left and right sections taper from a root chord of $0.16m$ to a tip chord of $0.12m$, each measuring 0.22 m in width. The straight leading edge and tapered trailing edge create a root-to-tip sweep angle of $-2.291^{\circ}$. The wing’s aerodynamic center is located $0.0357m$ from the leading edge and the center of gravity (CG) is set $0.03m$ from the leading edge to ensure stability, with \(b\) and \(l\) denoting the distances from CG to the servo tilt axis along \(y_b\) and \(z_b\) axes. The aircraft weighs $489$ g including a $650$ mAh Li-Po battery and is equipped with a Pixracer R15 autopilot programmed running the Paparazzi open-source software \cite{baskaya2016flexible}, a pitot tube for airspeed measurement, an M9N-5883 GPS for localization, and a TFMini-S micro LiDAR module for height measurement during landing.

\subsection{Wind Tunnel Experiment}
\subsubsection{Tilt rotors}
The proposed tiltrotor-tailsitter achieves thrust vectoring by independently tilting the left and right rotors. The tilt mechanism is achieved with a customized 3D print design presented in Fig. \ref{fig:tilt}.

\subsubsection{Elevons}
Our previous research \cite{lovell2023attitude} identifies a limitation in roll moment control during forward flight when relying solely on thrust vectoring for control moment generation. When the rotors tilt upwards or downwards, the generated propwash alters the local angle of attack of the wing opposite to the thrust vector direction, potentially reducing the local lift and diminishing the roll moment produced by tilt rotors. 

To quantitatively address this, a new wind tunnel experiment has been performed to assess the control moment generation capabilities of rotor tilt and elevon deflection. As shown in Fig. \ref{fig:wind}, the experiment took place at TU Delft's Open Jet Facility wind tunnel, which measures 2.85 m in both width and height. The TRE-tailsitter was securely positioned in front of the wind tunnel's open test section, oriented directly into the wind tunnel airflow to create forward flight conditions.
The experimental setup measured forces and moments under different motor throttle settings (0\%, 30\%, 50\%, and 70\%) for different elevon deflections or rotor tilt angles (0, $\pm{18.9^{\circ}}$, $\pm{37.8^{\circ}}$, and $\pm{63^{\circ}}$) at zero angle of attack and $15$ m/s airspeed. Each throttle and elevon deflection/rotor tilt combination was measured for 10 seconds, and the mean values for each measurement were used for subsequent analysis.

Fig. \ref{fig:moment} compares the pitch and roll moments generated by elevon deflection and rotor tilt under various throttle settings. Fig. \ref{fig:moment} (left) shows that rotor tilt is significantly more effective at generating pitch moments than elevon deflection for high throttle settings (50\%, 70\%), while for relatively lower throttle settings (0\%, 30\%), elevons are capable of producing considerable pitch moment compared to tilt rotors. In Fig. \ref{fig:moment} (right) , it displays that elevon deflection is more effective at generating roll moments than rotor tilt for different throttle levels and tilt/deflection angles.
%, indicating that elevon deflection offers more control authority over roll moment generation in the forward flight phase.
Overall, Fig. \ref{fig:moment} highlights the strengths of each actuator, with rotor tilt inducing a stronger influence on pitch control and elevon deflection offering superior roll moment control.
The hybrid actuator configuration in the TRE-tailsitter combines the strengths of both types of actuators.
%, combining the agility and rapid response of thrust vectoring with the consistent control moments from elevon deflection, will offer sound and reliable control performance for a TRE-tailsitter.
\section{ATTITUDE CONTROL}
In this section, we address the solutions for the attitude control of the customized TRE-tailsitter, including the pivot takeoff/landing and in-flight phases.
\subsection{Pivot Takeoff and Landing}
In this paper, we propose a new takeoff and landing approach for a tailsitter aircraft through thrust vectoring, which lifts the drone off the ground by pivoting around its tail, eliminating the need of an upright stand before takeoff for conventional tailsitters. To enable controlled takeoff and landing in this pivoting configuration, we have developed a dedicated control strategy, detailed as follows.

\begin{figure}[!t]
	\centering
	\includegraphics[width=0.45\linewidth]{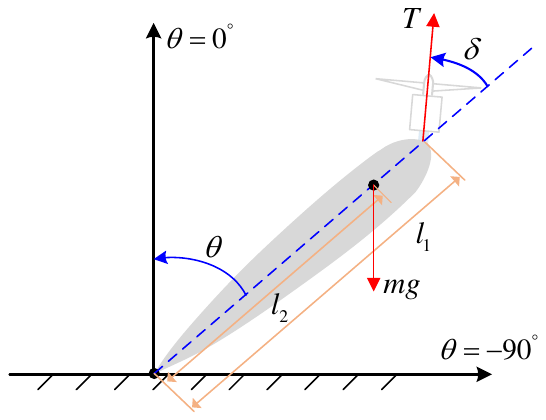}
	\caption{Schematic representation of the pivot control system.}
	\label{fig:pivot}
    \vspace{-10pt}
\end{figure}

Fig. \ref{fig:pivot} illustrates a schematic representation of the pivot control system. In this configuration, \( \theta \) represents the pitch angle, while \( I_{yy}^{\prime} \) denotes the moment of inertia about the tail pivot axis. The distances from the tilt servo axis and the CG to the tail pivot axis are denoted as \( l_1 \) and \( l_2 \) respectively. Both the left and right rotors are designed to have the same tilt angle and thrust command to prevent roll and yaw movements caused by differential thrust along the $z_b$ and $x_b$ axes. The primary control objective is to regulate the pitch angle \( \theta \) to track the desired pitch angle \( \theta_d \).

The longitudinal dynamics of the pivot system are described by the following equations of motion:
\begin{align}
	\begin{cases}
		I_{yy}^{\prime} \cdot \dot{q} = l_1 T \sin \delta - mg l_2 \sin(-\theta), \\
		\dot{\theta} = q,
	\end{cases}
	\label{eq:moment}
\end{align}
where \(m\) is the mass of the drone, $T$ is the combined thrust of both propellers, $\delta$ represents the tilt angle, 
%\( T \sin \delta \) represents the overall thrust component along the \( b_z \) axis, 
and \( \dot{q} \) is the angular acceleration around $y_b$ axis.

Next, we define the state and control input as:
\begin{align}
	x_1 = \theta - \theta_d, \quad x_2 = q, \quad u = T \sin \delta,
    \label{eq:statesinputs}
\end{align}
which leads to the following system model:
\begin{equation}
    \begin{cases}
        \dot{x}_1 = x_2 - \dot{\theta}_d, \\
        \dot{x}_2 = \frac{mg l_2 \sin \theta}{I'_{yy}} + \frac{l_1}{I'_{yy}} u.
    \end{cases}
\end{equation}
For Lyapunov stability analysis, we define an auxiliary state \( z = k_1 x_1 + x_2 - \dot{\theta}_d \), whose time derivative is
\begin{equation}
    \dot{z} = k_1 x_2 - k_1 \dot{\theta}_d
 + \frac{mg l_2 \sin \theta}{I'_{yy}} + \frac{l_1}{I'_{yy}} u -\ddot{\theta}_{d}, 
\end{equation}
where \( k_1 \) is a positive parameter. To stabilize the auxiliary state \( z \), we design the controller as follows:
\begin{equation}
    u = \frac{I'_{yy}}{l_1} \left( -k_1 x_2 
    + k_1 \dot{\theta}_d
    - \frac{mg l_2 \sin \theta}{I'_{yy}} +\ddot{\theta}_{d} 
    - k_2 z \right), \label{eq:expandedu}
\end{equation}
where \( k_2 > 0 \) is a control gain. To analyze the closed-loop stability, a Lyapunov function candidate is chosen as \( E = \frac{1}{2} z^2 \). With (1)-(4), we have
\begin{equation}
    \dot{E} = -k_2 z^2 = -2k_2 E.
\end{equation}
This negative semi-definite expression for \( \dot{E} \) ensures that the closed-loop system is globally asymptotically stable, which implies that \( z \to 0 \) as \( t \to +\infty \). When \( z = 0 \),
we have \( x_2 = \dot{\theta}_d - k_1 x_1 \Rightarrow \dot{x}_1 = -k_1 x_1 \). It follows that \( x_1(t) = x_1(0)e^{-k_1t} \), which implies the pitch error \( x_1 = \theta - \theta_d \) converges to zero.

Subsequently, the overall control input \( u \) needs to be allocated between thrust \( T \) and tilt angle \( \delta \). In real implementation, the desired pitch angle is set to be either a constant angle or a ramp variable, so \( \dot{\theta}_d \) is 0 or a constant, and \( \ddot{\theta}_d \) is 0. The auxilary state \( z \) into Eq. \eqref{eq:expandedu} yields:
\begin{equation}
    u = \underbrace{\frac{I_{yy}}{l_1} k_1 k_2 x_1 - \frac{I_{yy}}{l_1} (k_1 + k_2) (x_2-\dot{\theta}_d)}_{\text{\normalsize$\Delta u$}} \underbrace{-\frac{mg l_2 \sin \theta}{l_1}}_{\text{\normalsize$u_{eq}$}},
\end{equation}
where the control input is composed of two parts: the equilibrium component $u_{eq}$ and the feedback correction $\Delta u$ which refers to the real-time adjustments required to achieve the desired state. Specifically,
\begin{align}
	u_{eq} = - \frac{mg l_2 \sin \theta}{l_1},\quad  T_{eq} = \frac{mg l_2}{l_1}, \quad  \delta_{eq} = - \theta.
\end{align}
In order to allocate $\Delta u$ to $T$ and $\delta$, we first linearize $u$ in Eq. \eqref{eq:moment} around the equilibrium condition, so that we find the control effectiveness matrix \( B \) for thrust and tilt angle as follows:
\begin{align}
	B 
    %&= \begin{bmatrix}
	%	\frac{\partial M}{\partial T} & \frac{\partial M}{\partial {\delta}}
	%\end{bmatrix} 
    =  \begin{bmatrix}
		\sin \delta_{eq} & T_{eq} \cos \delta_{eq}
	\end{bmatrix}.
\end{align}
% The weighted pseudo inverse of \( B \) is then computed as:
% \begin{align}
% 	B^{+} &= W^{-1} B^T (B W^{-1} B^T)^{-1}.
% \end{align}
\begin{figure}[!t]
	\centering
	\includegraphics[width=0.28\textwidth]{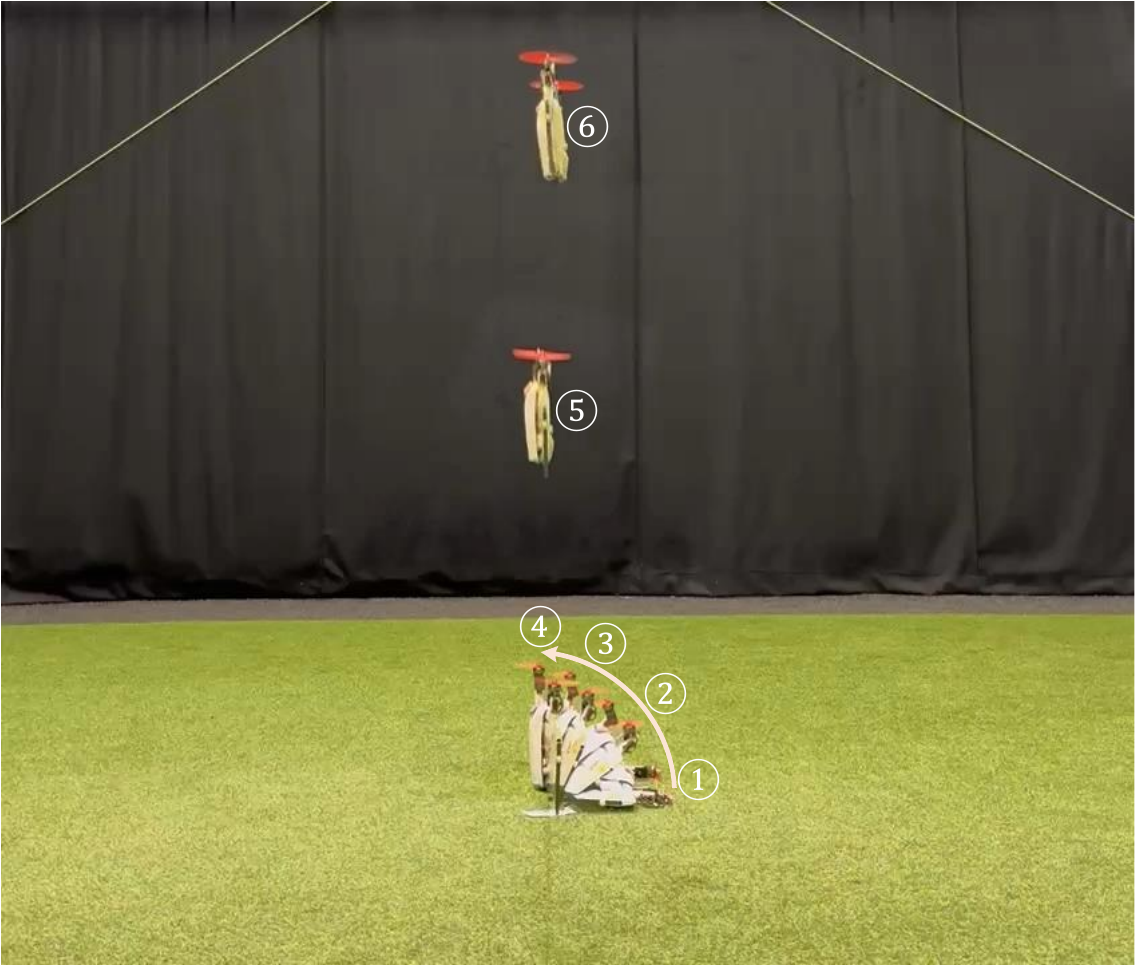}
	\caption{Indoor pivoting takeoff maneuver, captured at 0.8-second intervals of sequential video frames.}
	\label{fig:VTOL}
    \vspace{-10pt}
\end{figure}
The changes in thrust and tilt angle, $\Delta{T_{out}}$, $\Delta{\delta_{out}}$, are then calculated with the weighted pseudo inverse:
\begin{align}
	\begin{bmatrix}
		\Delta T, \Delta \delta
	\end{bmatrix} &= W^{-1} B^T (B W^{-1} B^T)^{-1} \Delta u,
\end{align}
where the weighting matrix compensates for the different scales of $T$ and $\delta$ and is given by:
\begin{align}
	W = \begin{bmatrix}
		\frac{1}{8.56^2} & 0 \\
		0 & \frac{1}{(63 \pi/180)^2}
	\end{bmatrix}.
\end{align}
The weight for thrust is empirically determined and the tilt angle weight corresponds to the servo's rotation range.
Finally, overall actuator commands of thrust and the tilt angle are given as:
\begin{align}
	T = \Delta_{out} + T_{eq} ,\quad \delta = \Delta \delta_{out} + \delta_{eq}.
\end{align}

By implementing the designed controller and control allocation method, the TRE-tailsitter is capable of VTOL in a pivoting pattern, as is shown in Fig. \ref{fig:VTOL}. With the drone lying on the ground, the rotors initially tilt up to their maximum angle, to initiate an upward thrust. Then the pivot controller governs the drone to pitch up by pivoting around its tail. Once the drone reaches a pitch angle of ±5.4° (6\% of 90°) and the pitch angular velocity falls below 5.73°/s (0.1 rad/s), indicating a stable upright state, the drone seamlessly enters the in-flight stage. For landing, the pivot controller is triggered again when the LiDAR sensor detects the altitude falling below 0.15 m and the tailsitter then pivots back to the ground in an inverse manner to takeoff.

\subsection{In-flight INDI Attitude Controller}
Due to the high non-linearity inherent in the dynamics of a TRE-tailsitter, an INDI controller is applied during flight, for its minimal dependency on the vehicle model \cite{smeur2016adaptive}.
Since the INDI controller relies on an estimate of the actuator state, the actuator dynamics are modeled as:
\begin{equation}
    A(s) = \frac{1}{\tau s + 1},
\end{equation}
where \(\tau_{servo} = 0.00325s\) with a rate limit of \(12.54\) rad/s; \(\tau_{motor} = 0.00707s\) with no rate limit. 

Let \(\boldsymbol{\Omega}\) denote the angular velocity vector of the vehicle, \(\boldsymbol{v}\) the velocity vector in the body frame, and \(\boldsymbol{T} = \left[T_l, T_r \right]^T\) the left and right motor thrusts, respectively. Assuming that changes in inertia due to the rotor tilts are negligible, and considering a diagonal inertia matrix \(\boldsymbol{I}\), Euler’s equation for rotational motion can be expressed as:
\begin{equation}
	\label{eq:eule_rota_equa}
	\boldsymbol{I \dot{\Omega}} + \boldsymbol{\Omega} \times \boldsymbol{I \Omega} = \boldsymbol{M}_{\delta_{T}}(\boldsymbol{T}, \boldsymbol{\delta}_{T}) + \boldsymbol{M}_{a}(\boldsymbol{\Omega}, \boldsymbol{v},\boldsymbol{\delta}_{E} ),
\end{equation}
where \(\boldsymbol{M}_{a}\) denotes the aerodynamic moments acting on the vehicle, and \(\boldsymbol{M}_{\delta_{T}}\) represents the control moments generated by the rotor tilt which can be further expanded as:
\begin{equation}
	\label{eq:m_c_appe}
	\boldsymbol{M}_{\delta_{T}} =\left[\begin{array}{c}
		b T_{L} \cos{\delta_{T_L}} - b T_{R} \cos{\delta_{T_R}}\\
		\,\,\,l T_{L} \sin{\delta_{T_L}} \,\,+ l T_{R} \sin{\delta_{T_R}}\\
		\,\,\,-b T_{L} \sin{\delta_{T_L}} \,\,+ b T_{R} \sin{\delta_{T_R}}
	\end{array}\right],
\end{equation}
The angular accelerations of the vehicle are obtained by rearranging Euler’s equation as:
\begin{equation}
	\label{eq:eule_rota_equa_2}
	\boldsymbol{\dot{\Omega}} = \boldsymbol{I}^{-1}\left(  \boldsymbol{M}_{\delta_{T}}(\boldsymbol{T}, \boldsymbol{\delta}_T)+ \boldsymbol{M}_{a}(\boldsymbol{\Omega}, \boldsymbol{v}, \boldsymbol{\delta}_E)\right),
\end{equation}
% The angular acceleration is updated incrementally, and its fir-order Taylor expansion can be written as:
% \begin{equation}
	% \renewcommand{\arraystretch}{0.8}
	% \label{eq:fo_tayl_expa_atti_cont}
	% \begin{split}
		% \bs{\dot{\Omega}} = \bs{\dot{\Omega}}_{0} 
		% &+ \left. \frac{\partial}{\partial \bs{\omega}}\left(\bs{I}^{-1}\bs{M}_{T}(\bs{\omega}, \bs{\delta}_{T_0})\right)\right|_{\bs{\omega}=\bs{\omega}_{0}}\!\!\!\! \left(\bs{\omega}-\bs{\omega}_{0}\right) \\
		% &+ \left. \frac{\partial}{\partial \bs{\delta}}\left(\bs{I}^{-1}\bs{M}_{T}(\bs{\omega}_{0}, \bs{\delta}_{T})\right)\right|_{\bs{\delta}_{T}=\bs{\delta}_{T_0}}\!\!\!\! \left(\bs{\delta}_{T}-\bs{\delta}_{T_0}\right)\\
		% &+ \left. \frac{\partial}{\partial \bs{\Omega}}\left(\bs{I}^{-1}\bs{M}_{a}(\bs{\Omega}, \bs{v}_0, \bs{\delta}_{F_0})\right)\right|_{\bs{\Omega}=\bs{\Omega}_{0}}\!\!\!\! \left(\bs{\Omega}-\bs{\Omega}_{0}\right) \\
		% &+ \left. \frac{\partial}{\partial \bs{v}}\left(\bs{I}^{-1}\bs{M}_{a}(\bs{\Omega}_0, \bs{v}, \bs{\delta}_{F_0})\right)\right|_{\bs{v}=\bs{v}_{0}}\!\!\!\! \left(\bs{v}-\bs{v}_{0}\right)\\
		% &+ \left. \frac{\partial}{\partial \bs{\delta}_{F}}\left(\bs{I}^{-1}\bs{M}_{a}(\bs{\Omega}_0, \bs{v}_0, \bs{\delta}_{F})\right)\right|_{\bs{\delta}_{F}=\bs{\delta}_{F_0}}\!\!\!\! \left(\bs{\delta}_{F}-\bs{\delta}_{F_0}\right).
		% \end{split}
	% \renewcommand{\arraystretch}{1.0}
	% \end{equation}
where we assume the drone rotates slowly enough that the term $\bs{\Omega} \times \bs{I} \bs{\Omega}$ is negligible compared to the other terms. To incorporate thrust control along the negative \(z_b\) axis, the desired overall trust \(T_z\) is expressed as:
\begin{equation}
	T_Z = \frac{1}{m} \left( T_L \cos \delta_{T_L} + T_R \cos \delta_{T_R} \right),
	\label{eq:tz}
\end{equation}
The system dynamics consisting of the drone's angular acceleration and overall thrust can be expressed as:
\begin{equation}
	\label{eq:tayl_expa_indi_angu_acce_thru}
	\left[ \begin{array}{c}
		\boldsymbol{\dot{\Omega}} \\
		T_{Z}
	\end{array} \right] = \left[ \begin{array}{c}
		\boldsymbol{\dot{\Omega}}_{0}  \\
		T_{Z_{0}}
	\end{array} \right] + \boldsymbol{G}(\boldsymbol{u}-\boldsymbol{u}_{0}),
\end{equation}
where \( \boldsymbol{u} = \left[ \delta_{T_L}, \delta_{T_R}, T_{L}, T_{R}, \delta_{E_L}, \delta_{E_R} \right]^{T} \) represents a new command, and $\boldsymbol{u}_0$ represents the current actuator state that can be modeled from the control input vector \( \boldsymbol{u_c} \) with the actuator dynamics.
Then, the control effectiveness matrix \(\boldsymbol{G}\) is expressed as:
\begin{equation}
	\boldsymbol{G} = \left[\boldsymbol{G}_{\delta_{T}} \quad \boldsymbol{G}_{T} \quad \boldsymbol{G}_E \right],
\end{equation}
where \(\boldsymbol{G}_{\delta_{T}}\), \(\boldsymbol{G}_{T}\), and \(\boldsymbol{G}_E\) represent the control effectiveness of the rotor tilt, motor rotational speed and elevon deflection, respectively. Specifically,
\begin{equation}
	\renewcommand{\arraystretch}{1.2}
	\boldsymbol{G}_{\delta_{T}} =
	\begin{bmatrix}
		\frac{\partial (\boldsymbol{I}^{-1}\boldsymbol{M}_{\delta_{T}})}{\partial \delta_{T}}\\
		\frac{\partial T_Z}{\partial \delta_{T}}
	\end{bmatrix},
    \boldsymbol{G}_{T} =
	\begin{bmatrix}
		\frac{\partial (\boldsymbol{I}^{-1}\boldsymbol{M}_{T})}{\partial T}\\
		\frac{\partial T_Z}{\partial T}
	\end{bmatrix},
	\renewcommand{\arraystretch}{1.0}
\end{equation}
% \begin{equation}
	% \renewcommand{\arraystretch}{1.1}
	% \boldsymbol{G}_T =
	% \begin{bmatrix}
		% -bT_L \sin \delta_{TL} & bT_R \sin \delta_{TR} \\
		% lT_L \cos \delta_{TL} & lT_R \cos \delta_{TR} \\
		% -bT_L \cos \delta_{TL} & bT_R \cos \delta_{TR} \\
		% T_L \sin \delta_{TL} & T_R \sin \delta_{TR}
		% \end{bmatrix},
	% \renewcommand{\arraystretch}{1.0}
	% \end{equation}
% \begin{equation}
	% \renewcommand{\arraystretch}{1.5}
	% \boldsymbol{G}_{\omega} =
	% \begin{bmatrix}
		% b\cos\delta_{TL}\frac{\partial T_L}{\partial \omega_{L}} & b\cos\delta_{TR}\frac{\partial T_R}{\partial \omega_{R}} \\
		% l\sin\delta_{TL}\frac{\partial T_L}{\partial \omega_{L}} & l\sin\delta_{TR}\frac{\partial T_R}{\partial \omega_{R}} \\
		% -b\sin\delta_{TL}\frac{\partial T_L}{\partial \omega_{L}} & b\sin\delta_{TR}\frac{\partial T_R}{\partial \omega_{R}} \\
		% \cos\delta_{TL}\frac{\partial T_L}{\partial \omega_{L}} & \cos\delta_{TR}\frac{\partial T_R}{\partial \omega_{R}} 
		% \end{bmatrix},
	% \renewcommand{\arraystretch}{1.0}
	% \end{equation}
while \(\boldsymbol{G}_E\) accounts for the aerodynamic surfaces:
\begin{equation}
	\boldsymbol{G}_E =
	\begin{bmatrix}
		0 & 0 \\
		G_{E_{25}} & G_{E_{26}} \\
		G_{E_{35}} & G_{E_{36}} \\
		0 & 0
	\end{bmatrix}.
\end{equation}
Given the elevons are aerodynamic control surfaces, their control effectiveness is continuously affected by the airspeed \(V\).
To account for this, we schedule $\boldsymbol{G}_E$ based on the airspeed, as measured by a pitot tube. Since the pitot tube readings are inaccurate at low airspeed, the control effectiveness of elevon deflection is scheduled with the pitch angle at low airspeed. The functions of elevon control effectiveness, in units of (rad$/$s$^2$)$/$rad, are derived from flight test data. Specifically, the elevon control effectiveness about the pitch axis is given as:
\begin{equation} 
	\scalebox{0.9}{$
		G_{E_{25}}(\theta, V) =
		\begin{cases}
			13.10(1 - r_{\theta}) + 21.83 r_{\theta}, & V < 12 \, \text{m/s}, \\
			13.10 + 0.1746V^2, & V \geq 12 \, \text{m/s},
		\end{cases}
		$}
\end{equation}
% \begin{equation}
% 	\scalebox{0.9}{$
% 		G_{F_{24}}(\theta, V) =
% 		\begin{cases}
% 			(1.5(1 - r_{\theta}) - 2.5 r_{\theta}) \cdot 10^{-3}, & V < 12 \, \text{m/s}, \\
% 			(1.5 - 0.02V^2) \cdot 10^{-3}, & V \geq 12 \, \text{m/s},
% 		\end{cases}
% 		$}
% \end{equation}
with \(G_{E_{26}}\) = \(G_{E_{25}}\) and where \(r_{\theta}\) serves as a pitch ratio relative to the vertical and forward flight phases:
\begin{equation}
	\renewcommand{\arraystretch}{1.2}
	r_{\theta} =
	\begin{cases}
		0, &  -\frac{\pi}{6} \leq \theta, \\
		\left(\theta + \frac{\pi}{6}\right)/(-\frac{\pi}{6}), &  -\frac{\pi}{3} \leq \theta\leq -\frac{\pi}{6}, \\
		1, & \theta \leq -\frac{\pi}{3},
	\end{cases}
	\renewcommand{\arraystretch}{1.0}
\end{equation}
Similarly, the yaw control effectiveness is computed by:
\begin{equation}
	\scalebox{0.9}{$
		G_{E_{35}}(\theta, V) =
		\begin{cases}
			15.72(1 - r_{\theta}) + 26.19 r_{\theta}, & V < 12 \, \text{m/s}, \\
			15.72 + 0.0873V^2, & V \geq 12 \, \text{m/s},
		\end{cases}
		$}
\end{equation}
% \begin{equation}
% 	\scalebox{0.9}{$
% 		G_{F_{34}}(\theta, V) =
% 		\begin{cases}
% 			(1.8(1 + r_{\theta}) - 3.0 r_{\theta}) \cdot 10^{-3}, & V < 12 \, \text{m/s}, \\
% 			(1.8 - 0.01V^2) \cdot 10^{-3}, & V \geq 12 \, \text{m/s},
% 		\end{cases}
% 		$}
% \end{equation}
with \(G_{E_{36}}\) = \(- G_{E_{35}}\).

From Eq. \ref{eq:tayl_expa_indi_angu_acce_thru}, the control input $u$ is found using weighted least squares (WLS) algorithm presented in \cite{Smeur2020}, with the cost function defined as:
\begin{equation}
	C(u) = \|W_u(\boldsymbol{u} - \boldsymbol{u}_p)\|^2 + \gamma \|W_v(\boldsymbol{G}\boldsymbol{u} - \boldsymbol{\nu})\|^2,
\end{equation}
where \(\boldsymbol{u}_p\) represents the preferred control inputs and \(\boldsymbol{\nu}\) is the control objective in terms of $\left[\dot{\boldsymbol{\Omega}} \ \ T_Z \right]^T$, acquired from the errors in attitude angles and angular rates with a linear controller.
\(\gamma\) is a scale factor of \(10^4\), and \(W_u\) and \(W_v\) are control input and control objective weighting matrices respectively. \(W_v\) is defined as \([10, 10, 0.1, 1.0]\), with the weight corresponding to the priority of the rotation about the \(x_b\), \(y_b\), \(z_b\) axes and thrust respectively.

\begin{figure}[ht]
	\centering		\includegraphics[width=0.9\linewidth]{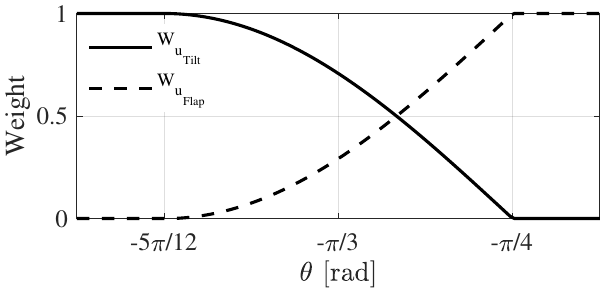}
	\caption{Weights for rotor tilt and elevon deflection.}
	\label{fig:weight}
\end{figure}

Since thrust vectoring is more effective in the vertical flight phase and elevon deflection is more stable and consistent especially for roll control in the forward flight phase, the controller is designed to prioritize rotor tilt in vertical flight and elevon deflection in forward flight. As is displayed in Fig. \ref{fig:weight}, the weights for rotor tilt and elevon deflection, bounded between 0.001 and 1, are adjusted dynamically and continuously based on the pitch angle, ensuring smooth actuator allocation across different flight phases. The computed weights apply to both left and right rotor tilts/elevon deflections. Throughout the flight envelope, the weights for motor thrust remain constant at 0.001.
\vspace{-0.4cm}
% \begin{equation}
% 	\renewcommand{\arraystretch}{1.2}
% 	W_{u_T} =
% 	\begin{cases}
% 		0.001, &  -\frac{\pi}{4} \leq \theta, \\
% 		\cos \left( 3 \cdot \theta +  \frac{5\pi}{4}  \right), &  -\frac{5 \pi}{12} \leq \theta\leq -\frac{\pi}{4}, \\
% 		1, & \theta \leq -\frac{5 \pi}{12}.
% 	\end{cases}
% 	\renewcommand{\arraystretch}{1.0}
% \end{equation}
% The weight for elevon deflection follows an opposite trend:
% \begin{equation}
% 	\renewcommand{\arraystretch}{1.2}
% 	W_{u_E} =
% 	\begin{cases}
% 		1, &  -\frac{\pi}{4} \leq \theta, \\
% 		-\cos \left( 3 \cdot \theta +  \frac{5\pi}{4}  \right)+1, &  -\frac{5 \pi}{12} \leq \theta\leq -\frac{\pi}{4}, \\
% 		0.001, & \theta \leq -\frac{5 \pi}{12},
% 	\end{cases}
% 	\renewcommand{\arraystretch}{1.0}
% \end{equation}
\begin{figure*}[h]
    \centering
    % Figure 9 (Left side)
    \begin{minipage}{0.4\textwidth}
        \centering
        \begin{subfigure}[t]{0.49\textwidth}
            \centering
            \includegraphics[height=4.2cm]{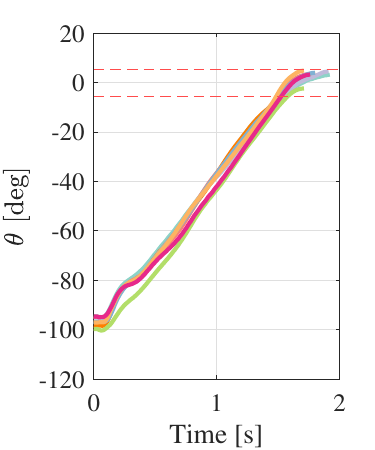}
            \caption{}
            \label{fig:pivot_theta}
        \end{subfigure}
        \hspace{-0.5cm}
        \begin{subfigure}[t]{0.49\textwidth}
            \centering
            \includegraphics[height=4.2cm]{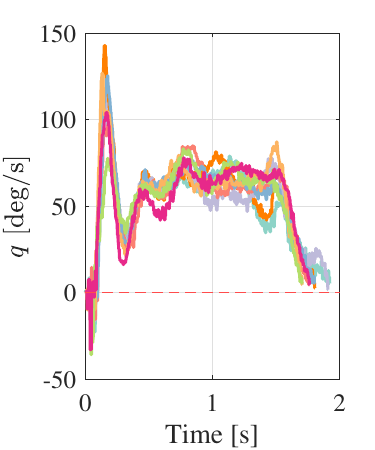}
            \caption{}
            \label{fig:pivot_rate}
        \end{subfigure}
        \caption{Eight sequent outdoor pivoting experiments. (a) Pitch angle. (b) Pitch rate.}
        \label{fig:pivoteight}
    \end{minipage}
    % Figure 10 (Right side)
    \begin{minipage}{0.58\textwidth}
        \centering
        \begin{subfigure}[t]{0.49\textwidth}
            \centering
            \includegraphics[width=\linewidth]{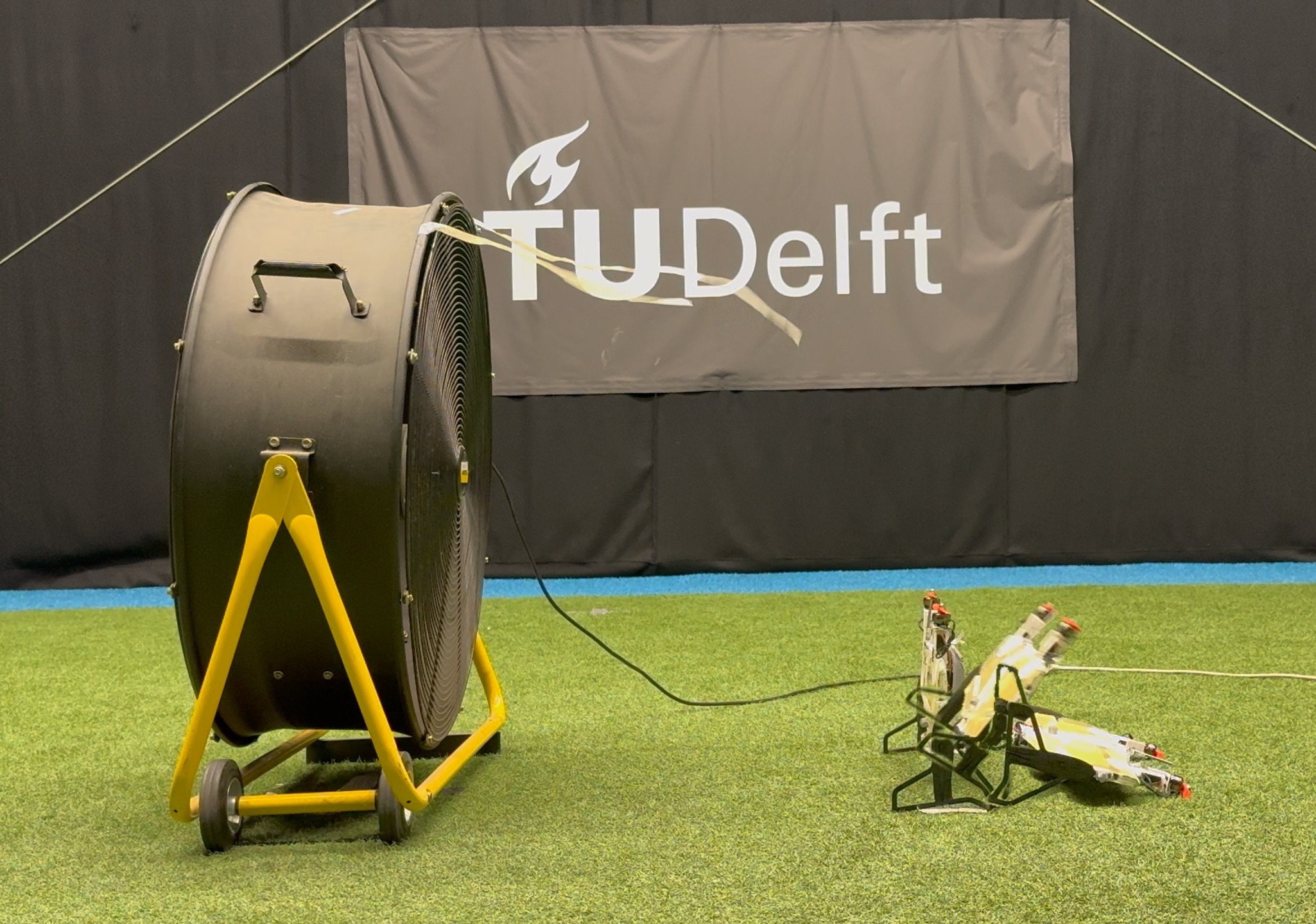}
            \caption{}
            \label{fig:high_E}
        \end{subfigure}
        \hspace{-0.2cm}
        \begin{subfigure}[t]{0.49\textwidth}
            \centering
            \includegraphics[width=\linewidth]{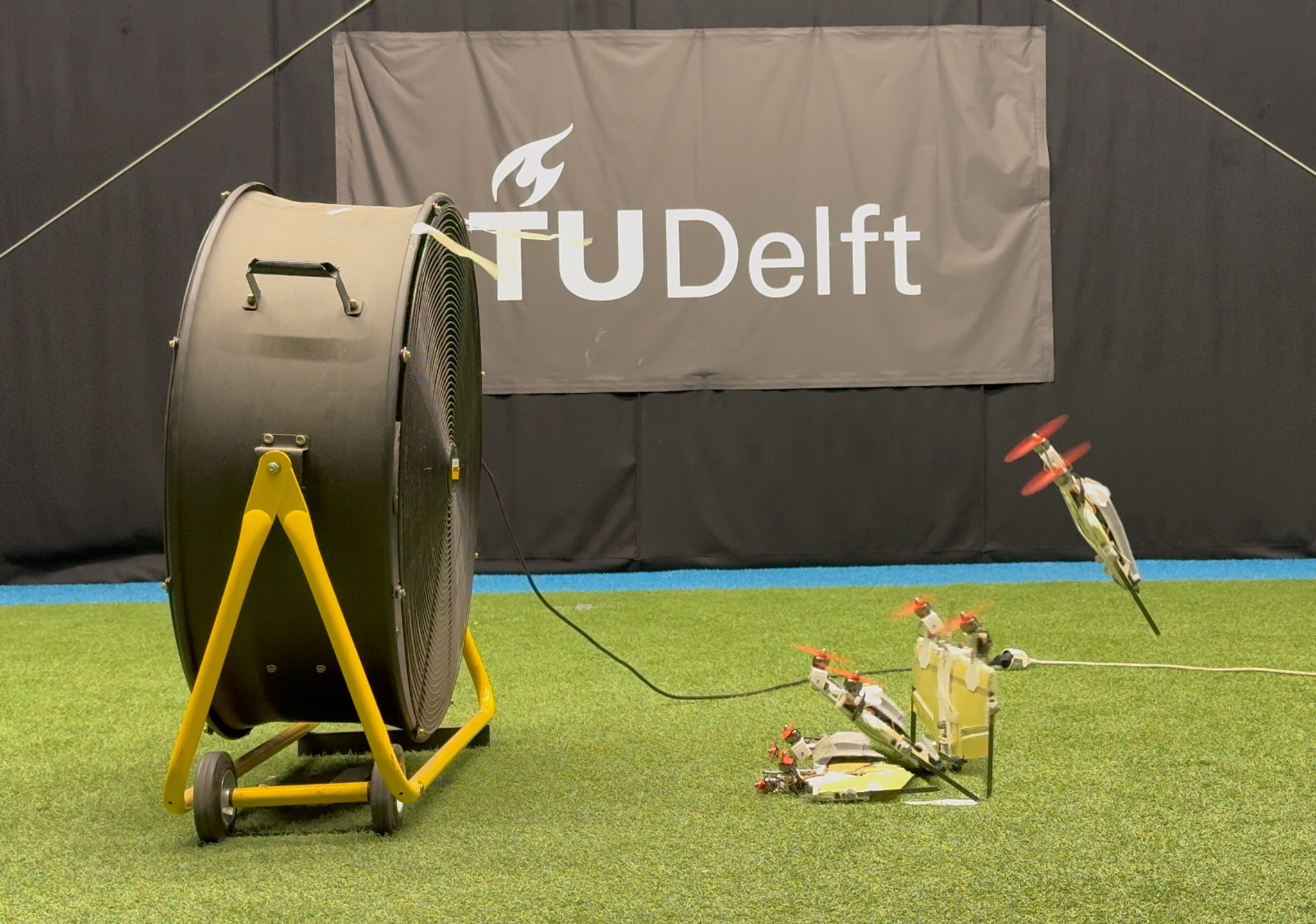}
            \caption{}
            \label{fig:high_TRE}
        \end{subfigure}
        \caption{Comparative tests between E-tailsitter and TRE-tailsitter under strong turbulent flow.}
        \label{fig:highwind}
    \end{minipage}
    \vspace{-18pt}
\end{figure*}
\section{VELOCITY CONTROL AND GUIDANCE}
The guidance of the TRE-tailsitter follows the implementation outlined in \cite{Smeur2020}. Similar to the INDI attitude controller, an outer loop WLS-based INDI controller is applied makin use of a control effectiveness matrix that maps changes in attitude angles and thrust to corresponding changes in linear accelerations. The heading of the tiltrotor-tailsitter is updated by a proportional sideslip correction controller, with the assumption that changes in heading are small relative to the other attitude angles and thrust. Therefore, the control input weight matrix, \(W_u = [1, 1, 1]\), assigns equal weights to roll, pitch, and thrust increments, while the control objective weight matrix, \(W_v = [100, 100, 1]\), prioritizes the horizontal linear accelerations in the North-East-Down (NED) frame. The reference linear acceleration is computed from position and velocity errors using a PD controller.

In this paper, the TRE-tailsitter is assigned an autonomous waypoint tracking flight plan. The reference path is defined as straight-line segments connecting the waypoints, and the velocity vector field method is used to guide the drone along the desired track.

\section{FLIGHT TEST RESULTS}
The customized TRE-tailsitter's flight performance was evaluated through outdoor tests with the implemented guidance and attitude controllers, and the flight test results are presented in this section.
% \begin{figure*}[h]
%     \centering
%     \begin{subfigure}[t]{0.2\textwidth}
%         \centering
%         \includegraphics[height= 4.2cm]{Figures/takeoff_theta2.eps}
%         \caption{}
%         \label{fig:wind1}
%     \end{subfigure}
%     \hspace{-0.5cm}
%     \begin{subfigure}[t]{0.2\textwidth}
%         \centering
%         \includegraphics[height= 4.2cm]{Figures/takeoff_q2.eps}
%         \caption{}
%         \label{fig:moment}
%     \end{subfigure}
%     \begin{subfigure}[t]{0.3\textwidth}
%         \centering
%         \includegraphics[width=\linewidth]{Figures/highwind_E.jpg}
%         \caption{}
%         \label{fig:wind2}
%     \end{subfigure}
%     \begin{subfigure}[t]{0.3\textwidth}
%         \centering
%         \includegraphics[width=\linewidth]{Figures/highwind_TRE.jpg}
%         \caption{}
%         \label{fig:wind3}
%     \end{subfigure}
%     \caption{Comparison of wind tunnel test setups and aerodynamic moment generation.}
%     \label{fig:windtunnel}
% \end{figure*}

\subsection{Pivoting robustness}
In order to validate the robustness of the pivoting controller, eight outdoor pivoting tests were conducted sequentially under an average wind of $6.7\,\text{m/s}$ and a peak gust of $10.28\,\text{m/s}$, with their pitch angles and pitch rates shown in Fig. \ref{fig:pivoteight}. For each test, the pitch angle reaches an allowable error margin of $\pm{5.4^{\circ}}$,  corresponding to $6\%$ of the desired movement range (from $-90^\circ$ to $0^\circ$) with the pitch rate below $\pm{5.73^{\circ}}/s$, suggesting a stable upright posture ready to enter the in-flight stage. To further demonstrate the benefit of the pivoting approach, a E-tailsitter and a TRE-tailsitter of the same airframe were placed close to a three-bladed fan, which generates strong turbulent flow. As Fig. \ref{fig:highwind} shows, the E-tailsitter tips over while the TRE-tailsitter steadily pitches up off the ground and flies against the wind, indicating the superior performance of pivoting takeoff under windy conditions.

	% \begin{figure}[h]
	% 	\centering
	% 	\begin{subfigure}{0.49\linewidth}
	% 		\centering
	% 		\includegraphics[height= 4.2cm]{Figures/takeoff_theta2.eps}
	% 		\caption{}
	% 		\label{fig:theta2}
	% 	\end{subfigure}
	% 	\hspace{-0.2cm}
	% 	\begin{subfigure}{0.49\linewidth}
	% 		\centering
	% 		\includegraphics[height= 4.2cm]{Figures/takeoff_q2.eps}
	% 		\caption{}
	% 		\label{fig:q2}
	% 	\end{subfigure}
	% 	\caption{Eight sequent outdoor pivoting experiments. (a) Pitch angle. (b) Pitch rate.}
	% 	\label{fig:takeoff_robust}
	% \end{figure}
    \begin{figure*}[tp]
	\centering
	\begin{subfigure}{0.6\linewidth}
		\centering
		\includegraphics[height=6cm]{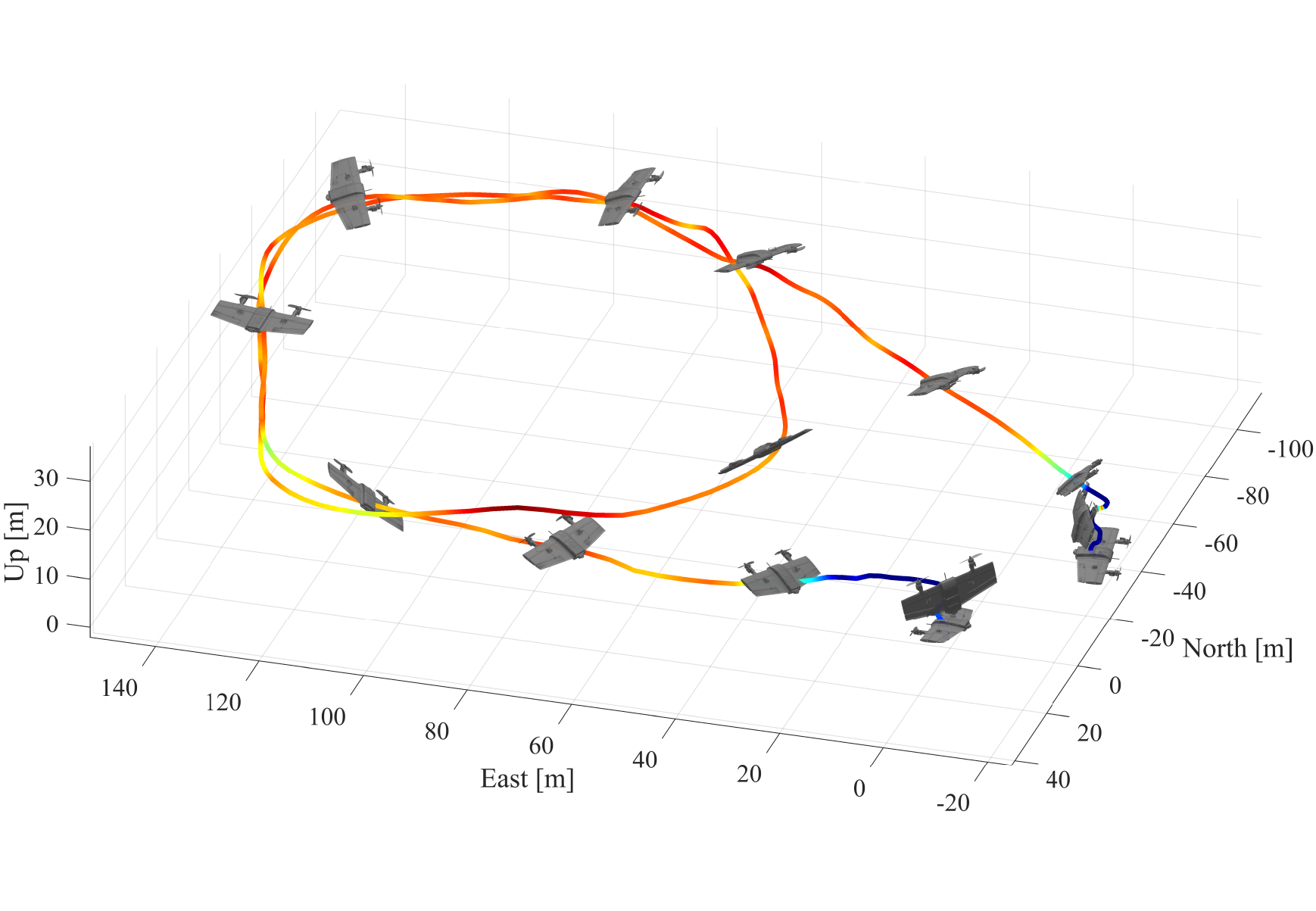}
		\vspace{-0.8cm}
		\caption{}
		\label{fig:3D}
	\end{subfigure}
	\hspace{-0.5cm}
	\raisebox{0.08cm}{
		\begin{subfigure}{0.4\linewidth}
			\centering
			\includegraphics[width=6.5cm]{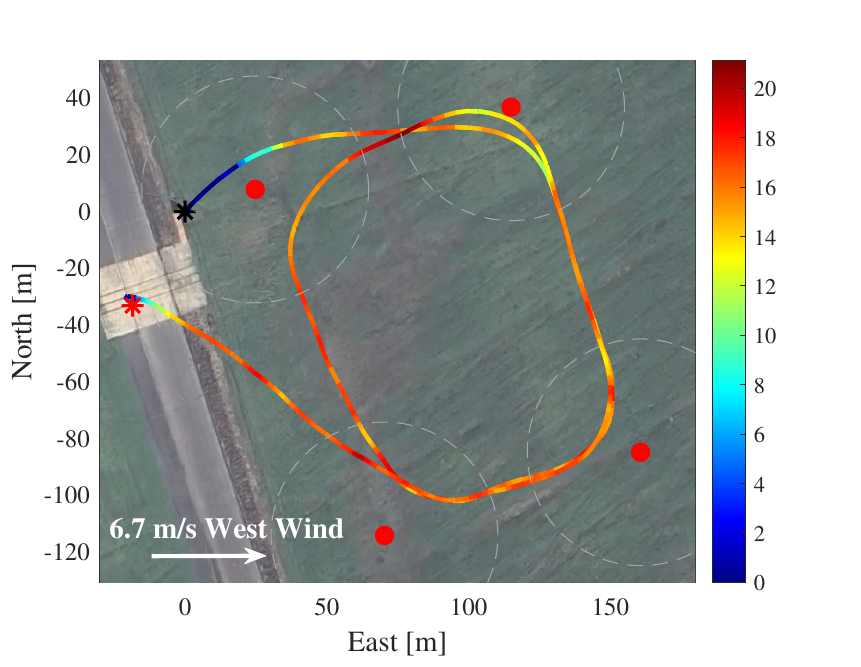}
			\caption{}
			\label{fig:2D}
		\end{subfigure}
	}
	\vspace{-15pt}
	\caption{ (a) 3D view of an autonomous, outdoor flight covering the full flight envelope. (b) 2D view of the waypoint tracking, with gray dashed lines representing the threshold range for reaching each waypoint.}
	\label{fig:track}
\end{figure*}

\begin{figure*}[t]
	\centering
	% First figure (Fig. 9)
	% \begin{minipage}{0.6\linewidth}
		\centering
		\begin{subfigure}{0.4\linewidth}
			\centering
			\includegraphics[width=\linewidth]{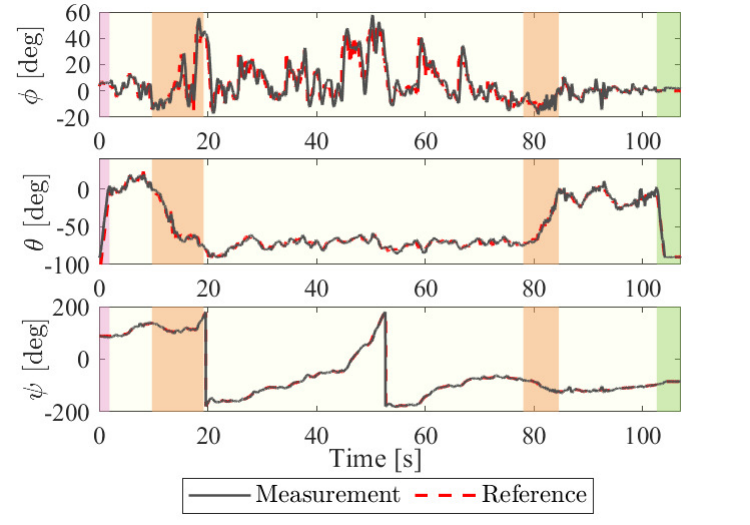}
			\caption{}
			\label{fig:attitude}
		\end{subfigure}
		\begin{subfigure}{0.4\linewidth}
			\centering
			\includegraphics[width=\linewidth]{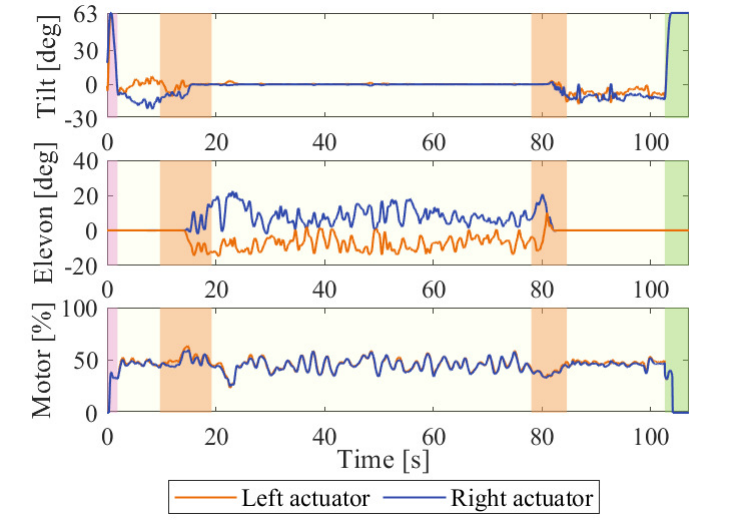}
			\caption{}
			\label{fig:actuator}
		\end{subfigure}
		\caption{(a) Measured and reference attitude angles corresponding to the autonomous flight shown in the 3D and 2D views. (b) Actuator states during the flight. Pink, light beige and green areas respectively refer to takeoff, in-flight and landing stages, with soft orange highlighting the transition phases.}
		\label{fig:fig9}
	% \end{minipage}
	% Second figure (Fig. 10)
	% \begin{minipage}{0.39\linewidth}
	% 	\centering
	% 	\begin{subfigure}{0.49\linewidth}
	% 		\centering
	% 		\includegraphics[height= 4cm]{Figures/takeoff_theta2.eps}
	% 		\caption{}
	% 		\label{fig:theta2}
	% 	\end{subfigure}
	% 	\hspace{-0.2cm}
	% 	\begin{subfigure}{0.49\linewidth}
	% 		\centering
	% 		\includegraphics[height= 4cm]{Figures/takeoff_q2.eps}
	% 		\caption{}
	% 		\label{fig:q2}
	% 	\end{subfigure}
	% 	\caption{Eight sequent pivot-up outdoor experiments. (a) Pitch angle. (b) Pitch rate.}
	% 	\label{fig:takeoff_robust}
	% \end{minipage}
    \vspace{-10pt}
\end{figure*}

\subsection{Autonomous full-envelope outdoor flight}
Fig. \ref{fig:3D} illustrates a 3D view of the TRE-tailsitter's trajectory throughout the entire flight envelope, demonstrating its capability to perform vertical takeoff and landing, and smooth transitions between vertical and forward flight phases. The flight begins with a pivoting vertical takeoff, followed by a seamless transition into forward flight. The drone autonomously tracks a series of waypoints before transitioning back to a controlled pivoting landing.
Fig. \ref{fig:2D} provides a top-down view of the waypoint tracking trajectory with color-coded airspeed. Despite a \(6.7\) m/s west wind and a peak gust of $10.28\,\text{m/s}$, the drone effectively reaches each waypoint at a cruise airspeed of \(16\) m/s, showcasing robust performance under wind disturbances.
% \begin{figure*}[h]
% 	% \vspace*{\fill}
% 	\centering
% 	\begin{subfigure}{0.3\linewidth}
% 		\centering
% 		\includegraphics[height= 4cm]{Figures/attitude.pdf}
% 		\caption{}
% 		\label{fig:attitude}
% 	\end{subfigure}
% 	\hspace{-0.4cm}
% 	\begin{subfigure}{0.3\linewidth}
% 		\centering
% 		\includegraphics[height= 4cm]{Figures/actuator.pdf}
% 		\caption{}
% 		\label{fig:actuator}
% 	\end{subfigure}
% 	\begin{subfigure}{0.19\linewidth}
% 		\centering
% 		\includegraphics[width=\linewidth]{Figures/takeoff_theta2.eps}
% 		\caption{}
% 		\label{fig:theta2}
% 	\end{subfigure}
% 	\hspace{-0.4cm}
% 	\begin{subfigure}{0.19\linewidth}
% 		\centering
% 		\includegraphics[width=\linewidth]{Figures/takeoff_q2.eps}
% 		\caption{}
% 		\label{fig:q2}
% 	\end{subfigure}
% 	\caption{a. First image description. b. Second image description.}
% \end{figure*}

Fig. \ref{fig:attitude} presents the measured and reference attitude angles (roll, pitch, and yaw) during the flight depicted in Fig. \ref{fig:track}. During vertical takeoff and landing, the drone exhibits a quick pitch change of approximately $\pm{90^{\circ}}$, indicating the pivoting takeoff. Smooth transitions are observed from the effective pitch angle changes. The close alignment between the measured and reference angles across all axes confirms stable control of the TRE-tailsitter across all flight phases. Correspondingly, Fig. \ref{fig:actuator} displays the actuator states during the flight, including rotor tilt angles, elevon deflections and motor throttle.
During takeoff, the rotors start from a maximum tilt angle and gradually adjust to a neutral position, and vice versa for the landing period. In flight, the rotor tilt is prioritized during vertical flight, while elevon deflection shows more prominent behavior in forward flight. Meanwhile, the motor speed remains relatively consistent throughout. Notably, no actuator saturation occurs during the entire flight, suggesting effective control authority.
\section{TRE-TAILSITTER VS. E-TAILSITTER}
% \subsection{Pivoting robustness}
% In order to validate the robustness of the pivoting controller, eight outdoor pivoting tests were conducted sequentially, with their pitch angles and pitch rates shown in Fig. \ref{fig:takeoff_robust}. For each test, the pitch angle reaches the expected range of $\pm{15^{\circ}}$ with the pitch rate below $\pm{5.73^{\circ}}/s$, suggesting the robustness and stability of the pivoting controller under wind disturbances.
% \subsection{Comparative tests}
To further investigate the flight performance of the TRE-tailsitter, comparative flight tests were conducted between the TRE-tailsitter and the E-tailsitter. By setting the weight for rotor tilt to a very high value (100000) and the weight for elevons to 0, the controller uses the elevons only. Additionally, with the actuator command for tilt set as 0, the TRE-tailsitter can be transformed into a E-tailsitter without any hardware modifications.

\begin{figure*}[h]
	\centering
	% First image (tilt_climb)
    
    \begin{subfigure}{0.32\linewidth}
		\centering
		\includegraphics[height = 4.8cm]{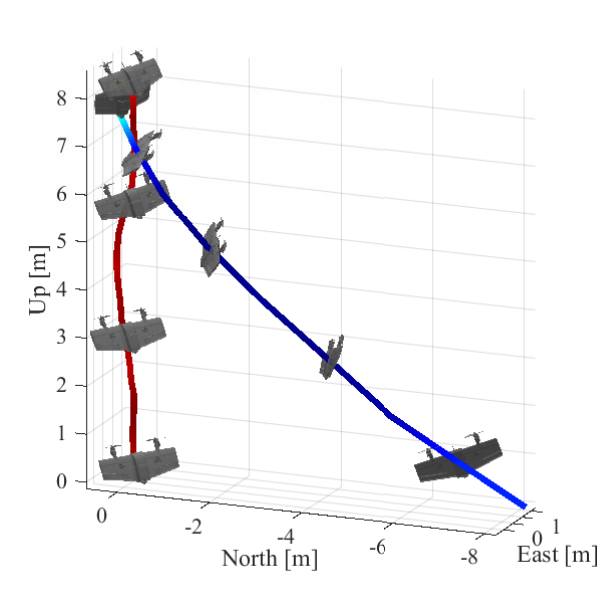}
     \hspace{-0.6cm} 
     \raisebox{0.5cm}{
   \includegraphics[height = 4cm]{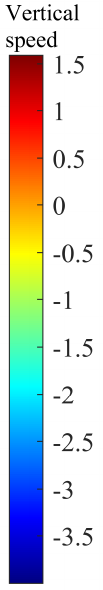}
   }
		\caption{E-tailsitter}
		\label{fig:flap_climb}
	\end{subfigure}
    \hspace{-0.4cm} 
	\begin{subfigure}{0.25\linewidth}
		\centering
		\includegraphics[height = 4.25cm]{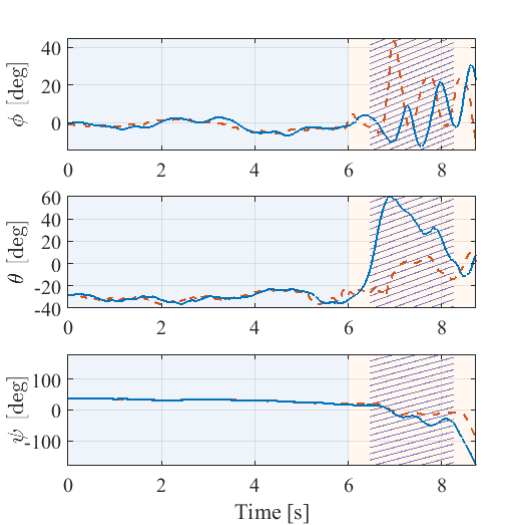}
  \includegraphics[width = 3.0cm]{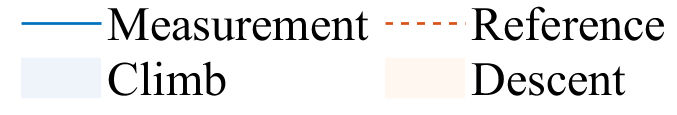}
		\caption{E-tailsitter attitude}
		\label{fig:flap_climb_att}
	\end{subfigure}
     \hspace{-0.3cm}
	\begin{subfigure}{0.17\linewidth}
		\centering
		\includegraphics[height = 5.0cm]{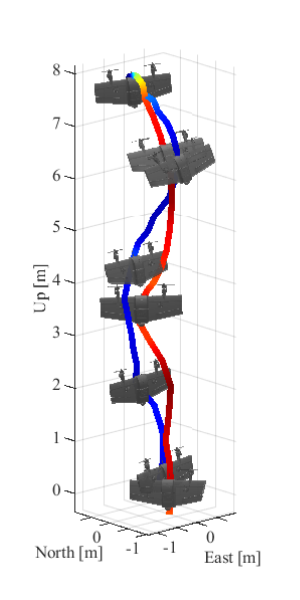}
   \hspace{-0.8cm} 
   \raisebox{0.8cm}{
   \includegraphics[height = 4cm]{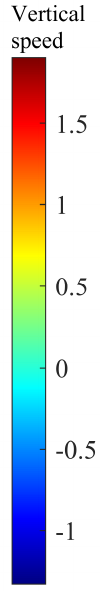}
   }
		\caption{TRE-tailsitter}
		\label{fig:tilt_climb}
	\end{subfigure}
      \hspace{-0.3cm} 
	\begin{subfigure}{0.25\linewidth}
		\centering
		\includegraphics[height = 4.25cm]{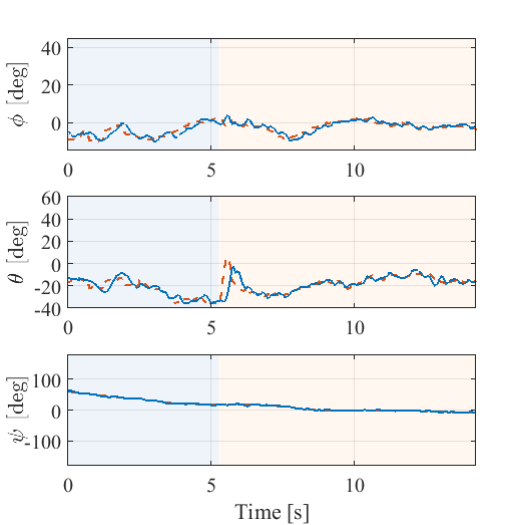}
  \includegraphics[width = 3.0cm]{Figures/legend_climb.png}
		\caption{TRE-tailsitter attitude}
		\label{fig:tilt_climb_att}
	\end{subfigure} 
	\caption{(a) 3D view of the climb and descent process for the E-tailsitter. (b) Attitude tracking for the E-tailsitter during its climb and descent, with shaded regions indicating at least one elevon saturation. (c) 3D view of the climb and descent process for the TRE-tailsitter. (d) Attitude tracking during the TRE-tailsitter's climb and descent. }
	\label{fig:climb_descent}
    \vspace{-15pt}
\end{figure*}

The first experiment is an 8-meter climb and descent test for both configurations under a 4 m/s north wind, in order to test the influence of vertical descending flight, in which the elevons may encounter reverse flow.
Fig. \ref{fig:climb_descent} presents the results.
During the descent phase, the E-tailsitter deviates significantly from the expected vertical path and the desired descent speed of $1m/s$, whereas the TRE-tailsitter maintains a stable climb and descent without notable position deviations. Quantitatively, the mean position error for the E-tailsitter is $0.291m$ during the climb and increases to 2.708 m during descent, compared to the TRE-tailsitter's lower mean errors of 0.256 m during climb and $0.348m$ during descent. Additionally, while no actuator saturation is observed for the TRE-tailsitter, the E-tailsitter encounters downward eleven saturation during descent. This demonstrates the improved flight control performance of the TRE-tailsitter, particularly during descent when the elevon control effectiveness is reduced due to insufficient or reversed airflow over the wings.

\begin{figure*}[tp]
	\centering
	% \hspace{-6cm} 
	% Group the first two images vertically
	\begin{subfigure}{0.32\linewidth}
		\centering
		% First image
		\hspace{-0.5cm}
		\includegraphics[width=\linewidth]{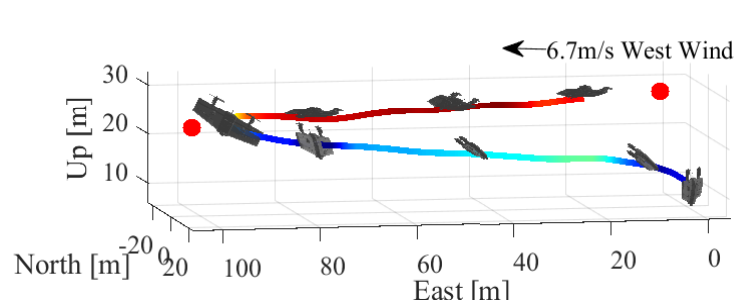}
		\caption{TRE-tailsitter}
		\label{fig:tran_tilt}
		
		\vspace{-0.05cm} % Adjust vertical space between first and second image
		
		% Second image
		\includegraphics[width=\linewidth]{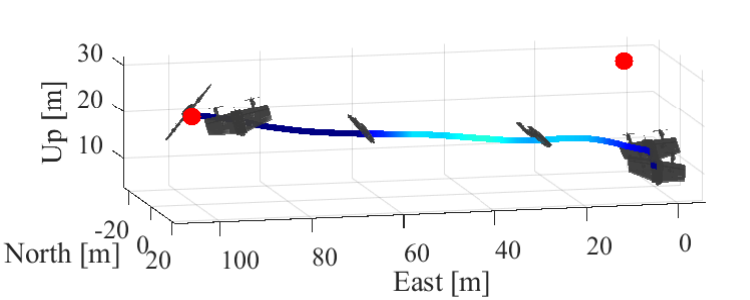}
		\caption{E-tailsitter}
		\label{fig:tran_flap}
	\end{subfigure}
 \hspace{-0.65cm} 
 \begin{subfigure}{0.033\linewidth}
		\centering
  \raisebox{0.9cm}{
		\includegraphics[width=\linewidth]{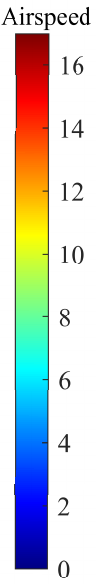}
}
		\label{fig:tran_airspeed}
	\end{subfigure}
	% Adjust horizontal space between groups
 \hspace{0.2cm} 
	% Third image to the right of the stacked images
	\begin{subfigure}{0.32\linewidth}
		\centering
		\includegraphics[width=5.1cm]{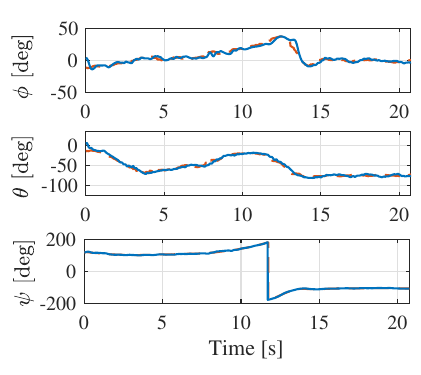}
        \includegraphics[width=4.5cm]{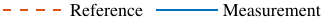}
		\caption{TRE-tailsitter attitude}
		\label{fig:tran_att_tilt}
	\end{subfigure}
  \hspace{-0.3cm} 
	\begin{subfigure}{0.32\linewidth}
		\centering
		\includegraphics[width=4.9cm]{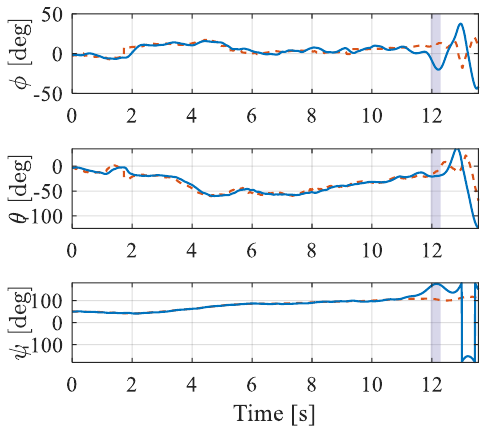}
   \includegraphics[width=4.5cm]{Figures/legend_turn_att.pdf}
		\caption{E-tailsitter attitude}
		\label{fig:tran_att_flap}
	\end{subfigure}
	\caption{(a-b) Transition and sharp turn process for the TRE-tailsitter and the E-tailsitter. (c) Attitude tracking for the TRE-tailsitter. (d) Attitude tracking for the E-tailsitter, with shaded regions indicating at least one elevon saturation.}
	\label{fig:sharp_turn}
    \vspace{-15pt}
\end{figure*}

In addition to vertical flight, comparative tests were conducted during the transition phase. As shown in Fig. \ref{fig:sharp_turn}, despite a 6.7 m/s westward wind, the TRE-tailsitter completes a sharp ''stop and go'' with an apex angle of $12.3^{\circ}$ with smooth transitions. In contrast, the E-tailsitter fails due to elevon saturation. Moreover, during the transition from hover to forward flight, the TRE-tailsitter predominantly relies on its tilt rotors, achieving a pitch control effectiveness of 126.6 \((rad/s^2)/rad\) compared to 27.1 \((rad/s^2)/rad\) for the elevon control effectiveness of the E-tailsitter. In the reverse transition, the TR-E tailsitter prioritizes elevons initially and then tilt rotors, where the pitch control effectiveness values for elevons and tilt rotors are respectively 37.9 \((rad/s^2)/rad\) and 98.6 \((rad/s^2)/rad\), versus 22.7 \((rad/s^2)/rad\) of that for E-tailsitter's elevons, highlighting the superiority of tilt rotors over elevon deflections during transitions. This comparative study underscores the TRE-tailsitter's capability of executing agile maneuvers, such as sharp turns, under wind disturbances.

% Time Range 1 (216.563s ~ 224.323s):
% MAE X: 0.119, MAE Y: 0.480
% RMSE X: 0.147, RMSE Y: 0.585
% Distance MAE: 0.509, Distance RMSE: 0.604
% Time Range 2 (472.063s ~ 479.163s):
% MAE X: 0.380, MAE Y: 0.437
% RMSE X: 0.467, RMSE Y: 0.522
% Distance MAE: 0.614, Distance RMSE: 0.701

% Time Range 1 (flap climb 3-19 m):
% MAE X: 0.198, MAE Y: 0.277
% RMSE X: 0.239, RMSE Y: 0.336
% Distance MAE: 0.366, Distance RMSE: 0.412
% Time Range 2 (flap descent 19-11m):
% MAE X: 2.699, MAE Y: 0.168
% RMSE X: 3.939, RMSE Y: 0.219
% Distance MAE: 2.708, Distance RMSE: 3.945

\section{Conclusion}
In this work, a TRE-tailsitter is designed and built to address the challenges of actuator saturation and limited control effectiveness for E-tailsitters during vertical flight and transitions. Additionally, the design overcomes the insufficient roll control authority observed in TR-tailsitters. The incorporation of rotor tilt not only enhances control but also introduces a novel pivoting takeoff and landing capability, eliminating the reliance on auxiliary supporting structures or landing gear commonly required by traditional E-tailsitters. The pivoting controller has demonstrated both stability and robustness under wind disturbances. A cascaded WLS-based INDI controller is implemented to maintain guidance and control of the drone across different flight phases. Wind tunnel tests validate the improved roll control authority of the TRE-tailsitter compared to TR-tailsitters, while comparative outdoor flight tests demonstrate its superior performance over E-tailsitters, particularly during vertical descent and transition phases. Moreover, the TRE-tailsitter exhibits high-speed forward flight capabilities like fixed-wing aircraft and agile maneuverability like quadrotors, particularly during sharp turns, indicating potential applications in drone racing. Future work will focus on exploring advanced trajectory planning algorithms to further optimize the drone's flight performance and agility.
%\section*{APPENDIX}
\section*{Acknowledgment}
The authors would like to thank Liming Zheng and Erik van der Host for their help with hardware implementations.

% \cite{yang2018active} proposes the active disturbance rejection attitude controller for outdoor hover and vertical flight. \cite{lyu2017hierarchical} achieves a full flight envelope for a quadrotor tail-sitter with a mode switch scheme between different flight phases. 
\balance
\bibliographystyle{IEEEtran}
\bibliography{IEEEabrv,IEEEexample}

\end{document}